\documentclass[runningheads]{llncs}
\usepackage{graphicx}
\usepackage[dvipsnames]{xcolor}
\usepackage{booktabs}
\usepackage{multirow}
\usepackage[export]{adjustbox}

\usepackage{multirow}
\usepackage[export]{adjustbox}

\begin{document}
\title{Multi-Task Bidirectional Transformer Representations for Irony Detection}%Multi-Task BERT for Arabic Irony Detection}

% 

%Limited Supervised Data for Irony Detection Bidirectional Encoder Representations from Transformers
\author{Chiyu Zhang \and
Muhammad Abdul-Mageed}
\authorrunning{Zhang and Abdul-Mageed}
\institute{Natural Language Processing Lab\\The University of British Columbia\\{\tt chiyuzh@mail.ubc.ca}, {\tt muhammad.mageeed@ubc.ca}}
\maketitle              % typeset the header of the contribution
\begin{abstract}
Supervised deep learning requires large amounts of training data. In the context of the FIRE2019 Arabic irony detection shared task (IDAT@FIRE2019), we show how we mitigate this need by fine-tuning the pre-trained bidirectional encoders from transformers (BERT) on gold data in a multi-task setting. We further improve our models by further pre-training BERT on `in-domain' data, thus alleviating an issue of dialect mismatch in the Google-released BERT model. Our best model acquires 82.4 macro \textit{$F_{1}$} score, and has the unique advantage of being feature-engineering free (i.e., based exclusively on deep learning). 

\end{abstract}

%-----------------------------------------------------------------------------------------------------------------------

\section{Introduction}
%---------------------
The proliferation of social media has provided a locus for use, and thereby collection, of figurative and creative language data, including irony ~\cite{ghosh2015semeval}. According to the Merriam-Webster online dictionary,~\footnote{\url{https://www.merriam-webster.com/dictionary/irony.}} \textit{irony} refers to ``the use of word to express something other than and especially the opposite of the literal meaning." A complex, controversial, and intriguing linguistic phenomenon, irony has been studied in disciplines such as linguistics, philosophy, and rhetoric. Irony detection also has implications for several NLP tasks such as sentiment analysis, hate speech detection, fake news detection, etc ~\cite{ghosh2015semeval}. Hence, automatic irony detection can potentially improve systems designed for each of these tasks. In this paper, we focus on learning irony. More specifically, we report our work submitted to the FIRE 2019 Arabic irony detection task (IDAT@FIRE2019).~\footnote{\url{https://www.irit.fr/IDAT2019/}} We focus our energy on an important angle of the problem--the small size of training data.

Deep learning is the most successful under supervised conditions with large amounts of training data (tens-to-hundreds of thousands of examples). For most real-world tasks, we hard to obtain labeled data. Hence, it is highly desirable to eliminate, or at least reduce, dependence on supervision. In NLP, pre-training language models on unlabeled data has emerged as a successful approach for improving model performance. In particular, the pre-trained multilingual \textbf{B}idirectional \textbf{E}ncoder \textbf{R}epresentations from \textbf{T}ransformers (BERT)~\cite{devlin2018bert} was introduced to learn language regularities from unlabeled data. Multi-task learning (MTL) is another approach that helps achieve inductive transfer between various tasks. More specifically, MTL leverages information from one or more source tasks to improve a target task~\cite{caruana1993,caruana1997multitask}. In this work, we introduce Transformer representations (BERT) in an MTL setting to address the data bottleneck in IDAT@FIRE2019. To show the utility of BERT, we compare to a simpler model with gated recurrent units (GRU) in a single task setting. To identify the utility, or lack thereof, of MTL BERT, we compare to a single task BERT model. For MTL BERT, we train on a number of tasks simultaneously. Tasks we train on are \textit{sentiment analysis}, \textit{gender detection}, \textit{age detection}, \textit{dialect identification}, and \textit{emotion detection}. 

Another problem we face is that the BERT model released by Google is trained only on Arabic Wikipedia, which is almost exclusively Modern Standard Arabic (MSA). This introduces a language variety mismatch due to the irony data involving a number of dialects that come from the Twitter domain. To mitigate this issue, we further pre-train BERT on an in-house dialectal Twitter dataset, showing the utility of this measure. To summarize, we make the following contributions:
\begin{itemize}
    \item In the context of the Arabic irony task, we show how a small-sized labeled data setting can be mitigated by training models in a multi-task learning setup.
    \item We view different varieties of Arabic as different domains, and hence introduce a simple, yet effective, `in-domain' training measure where we further pre-train BERT on a dataset closer to task domain (in that it involves dialectal tweet data).
\end{itemize}
%The rest of the paper is organized as follows: Section~\ref{methods} is about our methods. In Section~\ref{data}, we describe the IDAT@FIRE2019 dataset and the datasets we use for MTL. We introduce our models in Section~\ref{models}. Section~\ref{rel} is about related works, and Section~\ref{conc} is our conclusion.  

\section{Methods}~\label{methods}
%--------------------------------
\subsection{GRU}
For our baseline, we use gated recurrent units (GRU)~\cite{cho2014learning}, a simplification of long-short term memory (LSTM)~\cite{hochreiter1997long}, which in turn is a variation of recurrent neural networks (RNNs). A GRU learns based on the following:

%-----------------
\begin{equation}
 \textbf{\textit{h}}^{(t)} = \left( 1-\textbf{\textit{z}}^{(t)} \right) \textbf{\textit{h}}^{(t-1)} + \textbf{\textit{z}}^{(t)} \textbf{\textit{$\widetilde{h}$}}^{(t)} \\  
\end{equation}
%-------------------

where the \textit{update state} $\textbf{\textit{z}}^{(t)}$ decides how much the unit updates its content:

%-----------------
\begin{equation}
 \textbf{\textit{z}}^{(t)} = \sigma \left( W_z \textbf{\textit{x}}^{(t)} + U_z \textbf{\textit{h}}^{(t-1)} \right)   \\ 
\end{equation}
%-------------------

where W and U are weight matrices. The candidate activation makes use of a \textit{reset gate} $\textbf{\textit{r}}^{(t)}$: 
%-----------------
\begin{equation}
 \textbf{\textit{$\widetilde{h}$}}^{(t)} = tanh\left( W \textbf{\textit{x}}^{(t)} + \textbf{\textit{r}}^{(t)}  \odot \left(U \textbf{\textit{h}}^{(t-1)} \right)\right)   \\ 
\end{equation}
%-------------------

where $\odot$ is a Hadamard product (element-wise multiplication). When its value is close to zero, the reset gate allows the unit to \textit{forget} the previously computed state. The reset gate $\textbf{\textit{r}}^{(t)}$ is computed as follows:

%-----------------

\begin{equation}
 \textbf{\textit{$\widetilde{r}$}}^{(t)} = \sigma  \left( W_r \textbf{\textit{x}}^{(t)} + U_r \textbf{\textit{h}}^{(t-1)} \right)   \\ 
\end{equation}

%\textbf{Bidirectional Encoder Representations from Transformers (BERT).} 
\subsection{BERT} BERT~\cite{devlin2018bert} is based on the Transformer~\cite{vaswani2017attention}, a network architecture that depends solely on encoder-decoder attention. The Transformer attention employs a function operating on \textit{queries}, \textit{keys}, and \textit{values}. This attention function maps a query and a set of key-value pairs to an output, where the output is a weighted sum of the values. \textit{Encoder} of the Transformer in~\cite{vaswani2017attention} has 6 attention layers, each of which is composed of two sub-layers: (1) \textit{multi-head attention} where queries, keys, and values are projected \textit{h} times into linear, learned projections and ultimately concatenated; and 
%Intuitively, multi-head attention allows the model to jointly attend to information from different representation subspaces across different positions. 
(2) fully-connected \textit{feed-forward network (FFN)} that is applied to each position separately and identically. 
%The FFN has two linear layers and a ReLU activation function in-between. The FFN uses different parameters across the different layers. 
\textit{Decoder} of the Transformer also employs 6 identical layers, yet with an extra sub-layer that performs multi-head attention over the encoder stack. %Since the Transformer discards with both recurrence and convolution, it resorts to the so-called \textit{positional encoding} (based on sinusoidal wave functions) at the bottoms of the encoder and decoder stacks as a way to capture order of the sequence. %More information about the Transformer can be found in ~\cite{vaswani2017attention}. 
%As mentioned, the Transformer is the core learning component in BERT~\cite{devlin2018bert}, which we now introduce.
%As mentioned earlier, we use BERT to learn city, state, and country independently. 
% We implement and use the attention mechanism proposed by~\cite{vaswani2017attention}. The attention mechanism is based on 
The architecture of BERT~\cite{devlin2018bert} is a multi-layer bidirectional Transformer encoder \cite{vaswani2017attention}. It uses masked language models to enable pre-trained deep bidirectional representations, in addition to a binary \textit{next sentence prediction} task captures context (i.e., sentence relationships). More information about BERT can be found in~\cite{devlin2018bert}.

\subsection{Multi-task Learning} 
 In multi-task learning (MTL), a learner uses a number of (usually relevant) tasks to improve performance on a target task ~\cite{caruana1993,caruana1997multitask}. The MTL setup enables the learner to use cues from various tasks to improve the performance on the target task. MTL also usually helps regularize the model since the learner needs to find representations that are not specific to a single task, but rather more general. Supervised learning with deep neural networks requires large amounts of labeled data, which is not always available. By employing data from additional tasks, MTL thus practically augments training data to alleviate need for large labeled data. %Secondly, MTL learns universal representation across different tasks, which transfer learned knowledge from multiple domains and provide regularization effect via avoiding over-fitting to a specific task. 
 Many researchers achieve state-of-the-art results by employing MTL in supervised learning settings~\cite{guo2018soft,liu2019multi}. In specific, BERT was successfully used with MTL. Hence, we employ multi-task BERT (following~\cite{liu2019multi}). For our training, we use the same pre-trained BERT-Base Multilingual Cased model as the initial checkpoint. For this MTL pre-training of BERT, we use the same afore-mentioned single-task BERT parameters. We now describe our data.
 
%-------------------------------------------------------------

\section{Data}\label{data}

The shared task dataset contains 5,030 tweets related to different political issues and events in the Middle East taking place between 2011 and 2018. Tweets are collected using pre-defined keywords (i.e. targeted political figures or events) and the positive class involves ironic hashtags such as \#sokhria, \#tahakoum, and \#maskhara (Arabic variants for ``irony"). Duplicates, retweets, and non-intelligible tweets are removed by organizers. Tweets involve both MSA as well as dialects at various degrees of granularity such as \textit{Egyptian}, \textit{Gulf}, and \textit{Levantine}. 

IDAT@FIRE2019 \cite{idat2019} is set up as a binary classification task where tweets are assigned labels from the set \{\textit{ironic}, \textit{non-ironic}\}. A total of 4,024 tweets were released by organizers as training data. In addition, 1,006 tweets were used by organizers as test data. Test labels were not release; and teams were expected to submit the predictions produced by their systems on the test split. For our models, we split the 4,024 released training data into 90\% TRAIN ($n$=3,621 tweets; `ironic'=1,882 and `non-ironic'=1,739) and 10\% DEV ($n$=403 tweets; `ironic'=209 and `non-ironic'=194). We train our models on TRAIN, and evaluate on DEV. 

Our multi-task BERT models involve six different Arabic classification tasks. We briefly introduce the data for these tasks here:

\begin{itemize}
    \item \textbf{Author profiling and deception detection in Arabic (APDA).}~\cite{rangel2019ADPA}~\footnote{\url{https://www.autoritas.net/APDA/}}. From APDA, we only use the corpus of author profiling (which includes the three profiling tasks of \textit{age, gender}, and \textit{variety)}. The organizers of APDA provide 225,000 tweets as training data. Each tweet is labelled with three tags (one for each task). To develop our models, we split the training data into 90\% \textit{training} set ($n$=202,500 tweets) and 10\% \textit{development} set ($n$=22,500 tweets). With regard to age, authors consider tweets of three classes: \{\textit{Under 25}, \textit{Between 25 and 34}, and \textit{Above 35}\}. For the Arabic varieties, they consider the following fifteen classes: \{\textit{Algeria}, \textit{Egypt}, \textit{Iraq}, \textit{Kuwait}, \textit{Lebanon-Syria}, \textit{Lybia}, \textit{Morocco}, \textit{Oman}, \textit{Palestine-Jordan}, \textit{Qatar}, \textit{Saudi Arabia}, \textit{Sudan}, \textit{Tunisia}, \textit{UAE}, \textit{Yemen}\}. Gender is labeled as a binary task with \textit{\{male,female\}} tags.

\item \textbf{LAMA+DINA Emotion detection.} Alhuzali et al. \cite{alhuzali-etal-2018-enabling} introduce \textit{LAMA}, a dataset for Arabic emotion detection. They use a first-person seed phrase approach and extend work by Abdul-Mageed et al. ~\cite{mageed2016dina} for emotion data collection from 6 to 8 emotion categories (i.e. \textit{anger}, \textit{anticipation}, \textit{disgust}, \textit{fear}, \textit{joy}, \textit{sadness}, \textit{surprise} and \textit{trust}). We use the combined LAMA+DINA corpus. It is split by the authors as 189,902 tweets \textit{training} set, 910 as \textit{development}, and 941 as \textit{test}. In our experiment, we use only the training set for out MTL experiments. 

\item \textbf{Sentiment analysis in Arabic tweets.} This dataset is a shared task on Kaggle by Motaz Saad~\footnote{\url{https://www.kaggle.com/mksaad/arabic-sentiment-twitter-corpus}}. The corpus contains 58,751 Arabic tweets (46,940 \textit{training}, and 11,811 \textit{test}). The tweets are annotated with positive and negative labels based on an emoji lexicon.
 \end{itemize} 
%-------------------------------------------------------------
\section{Models}\label{models}
\subsection{GRU}\label{subsec:gru}
We train a baseline GRU network with our irony TRAIN data. This network has only one layer unidirectional GRU, with 500 unites and a linear, output layer. The input word tokens are embedded by the trainable word vectors which are initialized with a standard normal distribution, with $\mu=0$, and $\sigma=1$, i.e., $W \sim N(0,1)$. We use Adam \cite{kingma2014adam} with a fixed learning rate of $1e-3$ for optimization. For regularization, we use dropout~\cite{srivastava2014dropout} with a rate of 0.5 on the hidden layer. We set the maximum sequence sequence in our GRU model to 50 words, and use all 22,000 words of training set as vocabulary. We employ batch training with a batch size of 64 for this model. We run the network for 20 epochs and save the model at the end of each epoch, choosing the model that performs highest on DEV as our best model. We report our best result on DEV in Table ~\ref{tab:res}. Our best result is acquired with 12 epochs. As Table ~\ref{tab:res} shows, the baseline obtains \textit{$accuracy=73.70\%$} and \textit{$F_1=73.47$}.    
%-------------------------------------------------------------
\subsection{Single-Task BERT}\label{subsec:bert}
We use the BERT-Base Multilingual Cased model released by the authors~\cite{devlin2018bert}~\footnote{\url{https://github.com/google-research/bert/blob/master/multilingual.md}.}. The model is trained on 104 languages (including Arabic) with 12 layers, 768 hidden units each, 12 attention heads. The entire model has 110M parameters. The model has 119,547 shared WordPieces vocabulary, and was pre-trained on the entire Wikipedia for each language. For fine-tuning, we use a maximum sequence size of 50 tokens and a batch size of 32. We set the learning rate to $2e-5$ and train for 20 epochs. For single-task learning, we fine-tune BERT on the training set (i.e., TRAIN) of the irony task exclusively. We refer to this model as \textit{BERT-ST}, ST standing for `single task.' As Table ~\ref{tab:res} shows, BERT-ST unsurprisingly acquires better performance than the baseline GRU model. On accuracy, BERT-ST is 7.94\% better than the baseline. BERT-ST obtains 81.62 $F_1$ which is 7.35 better than the baseline.
%-------------------------------------------------------------

\subsection{Multi-Task BERT} \label{subsec:multi-bert}
We follow the work of Liu et al. \cite{liu2019multi} for training an MTL BERT in that we fine-tune the afore-mentioned BERT-Base Multilingual Cased model with different tasks jointly. First, we fine-tune with the three tasks of author profiling and the irony task simultaneously. We refer to this model trained on the 4 tasks simply as BERT-MT4. BERT-MT5 refers to the model fine-tuned on the 3 author profiling tasks, the emotion task, and the irony task. We also refer to the model fine-tuned on all six tasks (adding the sentiment task mentioned earlier) as BERT-MT6. For MTL BERT, we use the same parameters as the single task BERT listed in the previous sub-section (i.e., \textit{Single-Task BERT}). In Table ~\ref{tab:res}, we present the performance on the DEV set of only the irony detection task.~\footnote{We do not list acquired results on other tasks, since the focus of this paper is exclusively the IDAT@FIRE2019 shared task.} We note that all the results of multitask learning with BERT are better than those with the single task BERT. The model trained on all six tasks obtains the best result, which is 2.23\% accuracy and 2.25\% $F_1$ higher than the single task BERT model.

\begin{table}[]
\centering
\caption{Model Performance} \label{tab:res}
\begin{tabular}{@{}lcc@{}}
\toprule
\textbf{Model}         & \textbf{Acc} & \textbf{F1} \\ \midrule
\textbf{GRU}           & 0.7370       & 0.7347      \\ \midrule
\textbf{BERT-ST}   & \textbf{0.8164}       & \textbf{0.8162}      \\ \midrule
\textbf{BERT-MT4}      & 0.8189       & 0.8187      \\
\textbf{BERT-MT5}      & 0.8362       & 0.8359      \\
\textbf{BERT-MT6}      & \textbf{0.8387}       & \textbf{0.8387}      \\ \midrule
\textbf{BERT-1M-MT5} & \textbf{0.8437}       & \textbf{0.8434}      \\
\textbf{BERT-1M-MT6} & 0.8362       & 0.8360      \\ \bottomrule
\end{tabular}

\end{table}
%-------------------------------------------------------------
\subsection{In-Domain Pre-Training} 
Our irony data involves dialects such as Egyptian, Gulf, and Levantine, as we explained earlier. The BERT-Base Multilingual Cased model we used, however, was trained on Arabic Wikipedia, which is mostly MSA. We believe this dialect mismatch is sub-optimal. 
As Sun et al.~\cite{sun2019fine} show, further pre-training with domain specific data can improve performance of a learner. Viewing dialects as constituting different domains, we turn to dialectal data to further pre-train BERT. Namely, we use 1M tweets randomly sampled from an in-house Twitter dataset to resume pre-training BERT before we fine-tune on the irony data.~\footnote{A nuance is that we require each tweet in the 1M dataset to be $>20$ words long, and so this process is not entirely random.} We use BERT-Base Multilingual Cased model as an initial checkpoint and pre-train on this 1M dataset with a learning rate of $2e-5$, for 10 epochs. Then, we fine-tune on MT5 (and then on MT6) with the new \textit{further-pre-trained} BERT model. We refer to the new models as BERT-1M-MT5 and BERT-1M-MT6, respectively. As Table~\ref{tab:res} shows, BERT-1M-MT5 performs best: BERT-1M-MT5 obtains 84.37\% accuracy (0.5\% less than BERT-MT6) and 83.34 $F_1$ (0.47\% less than BERT-MT6). 

\subsection{IDAT@FIRE2019 Submission}
For the shared task submission, we use the predictions of BERT-1M-MT5 as our first submitted system. Then, we concatenate our DEV and TRAIN data to compose a new training set (thus using all the training data released by organizers) to re-train BERT-1M-MT5 and BERT-MT6 with the same parameters. We use the predictions of these two models as our second and third submissions. Our second submission obtains 82.4 $F_1$ on the official test set, and ranks $4th$ on this shared task.

%---------------
\section{Related Work}\label{rel}
%---------------------

\textbf{Multi-Task Learning.} MTL has been effectively used to model several NLP problems. These include, for example, syntactic parsing~\cite{luong2015multi}, sequence labeling~\cite{sogaard2016deep,rei2017semi}, and text classification~\cite{liu2016recurrent}. 

\textbf{Irony in different languages.} Irony detection has been investigated in various languages. For example, Hee et al.~\cite{van2018semeval} propose two irony detection tasks in English tweets. Task A is a binary classification task (\textit{irony} vs. \textit{non-irony}), and Task B is multi-class identification of a specific type of irony from the set \{\textit{verbal, situational, other-irony, non-ironic}\}. They use hashtags to automatically collect tweets that they manually annotate using a fine-grained annotation scheme. Participants in this competition construct models based on logistic regression and support vector machine (SVM)~\cite{rohanian2018wlv}, XGBoost ~\cite{rangwani2018nlprl}, convolutional neural networks (CNNs)~\cite{rangwani2018nlprl}, long short-term memory networks (LSTMs)~\cite{wu2018thu_ngn}, etc. For the Italian language, Cignarella et al. propose the IronTA shared task~\cite{cignarella2018overview}, and the best system~\cite{cimino2018multi} is a combination of bi-directional LSTMs, word $n$-grams, and affective lexicons. For Spanish, Ortega-Bueno1 et al.~\cite{ortega2019overview} introduce the IroSvA shared task, a binary classification task for tweets and news comments. The best-performing model on the task,~\cite{gonzalez2019elirf}, employs pre-trained Word2Vec, multi-head Transformer encoder and a global average pooling mechanism. 

\textbf{Irony in Arabic.} Arabic is a widely spoken collection of languages ($\sim$ 300 million native speakers)~\cite{mageedYouTweet2018,zhang2019no}. A large amount of works in Arabic are those focusing on other text classification tasks such as sentiment analysis~\cite{abdul2014samar,al2019comprehensive,abdul2017not,abdul2017modeling}, emotion~\cite{alhuzali-etal-2018-enabling}, and dialect identification~\cite{zhang2019no,elaraby2018deep,bouamor2018madar,bouamor2019madar}. Karoui et al.~\cite{karoui2017soukhria} created a Arabic irony detection corpus of 5,479 tweets. They use pre-defined hashtags to obtain irony tweets related to the US and Egyptian presidential elections. IDAT@FIRE2019~\cite{idat2019} aims at augmenting the corpus and enriching the topics, collecting more tweets within a wider region (the Middle East) and over a longer period (between 2011 and 2018).

%We now describe our methods. %the IDAT@FIRE2019 dataset.

%In this paper, we report our works on `Irony Detection in Arabic Tweets' shared task. We employ unidirectional Gated Recurrent Unit (GRU) \cite{cho2014learning} as our baseline and pre-trained multilingual Bidirectional Encoder Representations from Transformers (BERT) \cite{devlin2018bert} to identify irony for individual tweets. Due to the pre-trained multilingual BERT was trained with Wikipedia that mostly is MSA, we further pre-train BERT with a large amount of Arabic dialect corpus. Multi-task learning (MTL) is useful for learning multiple tasks jointly and transfer knowledge between different domain, which has been widely studied within NLP specifically. BERT with multi-task architecture has achieved state-of-the-art results on multiple natural language understanding (NLU) tasks \cite{liu2019multi}. Hence, we also employ the multi-task BERT model based on MT-DNN \cite{liu2019multi}. Our best system ranks 4th in the shared task. The rest of the paper is organized as follows: 
%-------------------------------------------------------------

\section{Conclusion}\label{conc}
In this paper, we described our submissions to the Irony Detection in Arabic shared task (IDAT@FIRE2019). We presented how we acquire effective models using pre-trained BERT in a multi-task learning setting. We also showed the utility of viewing different varieties of Arabic as different domains by reporting better performance with models pre-trained with dialectal data rather than exclusively on MSA. Our multi-task model with domain-specific BERT ranks $4th$ in the official IDAT@FIRE2019 evaluation. The model has the advantage of being exclusively based on deep learning. In the future, we will investigate other multi-task learning architectures, and extend our work with semi-supervised methods.

%-------------------------------------------------------------
\section{Acknowledgement}
We acknowledge the support of the Natural Sciences and Engineering Research Council of Canada (NSERC), the Social Sciences Research Council of Canada (SSHRC), and Compute Canada (\url{www.computecanada.ca}).
%-------------------------------------------------------------

%
%
%
\bibliographystyle{splncs04}
\bibliography{llncs.bib}
\end{document}